\documentclass[runningheads]{llncs}

\usepackage{eccv}

\usepackage{eccvabbrv}

\usepackage{graphicx}
\usepackage{booktabs}
\usepackage{wrapfig}
\usepackage{caption}

\usepackage[accsupp]{axessibility}  

\usepackage{hyperref}

\usepackage{orcidlink}

\begin{document}

\title{Distribution-Alignment Bridge for Uncertainty-Aware Text-to-Video Retrieval}

\titlerunning{Distribution-Alignment Bridge}

\author{Kyeongmo Chae\inst{1}\orcidlink{0009-0000-3504-4079} \and
  Jihoon Lee\inst{1}\orcidlink{0009-0001-6665-3739} \and
  Sangtae Ahn\inst{1}\thanks{Corresponding author: Sangtae Ahn (stahn@knu.ac.kr)}\orcidlink{0000-0001-9487-5649}}

\authorrunning{K.~Chae et al.}

\institute{School of Electronic and Electrical Engineering, Kyungpook National University, Daegu, South Korea 
  \\
  \email{\{ckm1994, leejh98123, stahn\}@knu.ac.kr}}

\maketitle

\begin{abstract}
  This paper proposes the Distribution-Alignment Bridge (DAB), a framework that reconceptualizes text-to-video retrieval as a distribution alignment task rather than traditional deterministic point matching. By modeling both text and video embeddings as Gaussian distributions defined by mean and variance, DAB explicitly accounts for modality-specific uncertainty. We employ a deterministic, diffusion-inspired bridge to iteratively refine text distributions toward their target video distributions through a truncated refinement process. This approach unifies probabilistic embedding and distributional transformation into a cohesive, end-to-end trainable system. To optimize cross-modal similarity, we introduce a distribution-aware contrastive loss based on Kullback–Leibler divergence. Extensive evaluations on MSR-VTT, MSVD, and VATEX benchmarks
  confirm that DAB significantly outperforms existing probabilistic and
  diffusion-based baselines, while providing calibrated uncertainty-aware
  ranking through bridge-induced distributional margins.
\end{abstract}

\section{Introduction}
\label{sec:intro}
Text-to-video retrieval (TVR), a pivotal task at the intersection of computer vision and natural language
processing, aims to identify the most semantically relevant video from a large corpus given a textual query.
The main challenge lies in the inherent modality gap, i.e., the representational disparity between the symbolic,
structured nature of text and the dense, spatio-temporal nature of video.
While recent methods leverage pretrained vision--language models such as CLIP~\cite{radford2021learning}
to project both modalities into a shared latent space, this modality gap persists and continues to hinder retrieval accuracy, especially under linguistic ambiguity and one-to-many correspondences.

A notable step forward is the Diffusion-Inspired Truncated Sampler (DITS)~\cite{li2023dits}, which reframes alignment as iterative refinement rather than one-shot projection. DITS demonstrated that directly applying generative diffusion models to retrieval suffers from instability: random noise initialization and \(\mathcal{L}_2\)-based objectives are ill-suited for ranking. Instead, DITS performs a deterministic truncated refinement that starts from meaningful text features and gradually evolves them toward target video embeddings under contrastive supervision, achieving both stability and efficiency.
However, despite these advances, DITS remains fundamentally point-wise: it treats text--video alignment as a mapping between single vectors and thus overlooks semantic uncertainty and one-to-many relations intrinsic to multimodal data. A single textual query can validly describe visually diverse scenes, and a single video can correspond to multiple semantically distinct captions. Collapsing such diversity into a single embedding point oversimplifies cross-modal semantics and restricts representational flexibility.

To overcome this limitation, we propose the Distribution-Alignment Bridge (DAB), a framework that casts alignment as a distribution-to-distribution transformation. Instead of deterministic vectors, DAB encodes both text and video as Gaussian distributions--the mean captures core semantics, while the variance reflects ambiguity and contextual diversity. At the heart of DAB is a Distribution Bridge, a diffusion-inspired yet deterministic transformation that iteratively refines the textual distribution toward the target video distribution through smooth, drift-guided updates.
Unlike generative diffusion models, the bridge is sampling-free and therefore stable and efficient for retrieval.
Specifically, DAB first extracts textual and frame-level video embeddings using CLIP, then aggregates frame features through a transformer-based cross-attention module to form text-conditioned video representations. A shared probabilistic encoder maps each modality to Gaussian parameters \((\mu, \log\sigma^2)\), explicitly separating semantic centers from uncertainty.
The bridge predicts drift terms \((\Delta\mu, \Delta\log\sigma^2)\) over truncated refinement steps to evolve the textual distribution toward its video counterpart. Crucially, DAB is trained with a distribution-aware contrastive objective,
in which the pairwise cost is a divergence between distributions--KL divergence by default (with \(W_2\) as a comparator)--
so that alignment becomes sensitive to both mean and variance, i.e., to information content as well as geometry.
This yields systematic distribution-space refinement of \((\mu,\log\sigma^2)\):
rather than mapping a query to a single point or simply reducing variance, DAB
learns a deterministic, sampling-free trajectory that jointly updates semantic
centers and uncertainty scales while preserving modality-specific uncertainty. Empirically, this not only improves top-\(k\) recall but also yields a substantial reduction in Mean Rank (MnR), demonstrating a globally more coherent embedding space rather than isolated top-rank gains.

The main contributions are as follows.
\begin{itemize}
  \item We identify the limitations of point-wise refinement (e.g., DITS) and
        introduce deterministic distribution-space refinement of \((\mu,\log\sigma^2)\),
        which jointly transports semantic centers and uncertainty scales without
        stochastic sampling.
  \item We formulate a bidirectional contrastive objective over directional KL costs,
        providing ranking-sensitive alignment of both mean and variance.
  \item Extensive experiments demonstrate state-of-the-art retrieval performance
        and show that DAB produces calibrated uncertainty signals and a semantically
        coherent embedding space.
\end{itemize}

\section{Related Work}

\subsection{Text-to-Video Retrieval}
Text-to-video retrieval (TVR) aims to find the most semantically relevant video for a given textual query.
Early studies~\cite{mithun2018learning,miech2019howto100m} learned a shared embedding space where similarity is computed using cosine or dot-product similarity.
With large-scale vision--language pre-training, CLIP-based frameworks such as CLIP4Clip~\cite{luo2022clip4clip} transferred image--text alignment knowledge to the video domain, achieving strong generalization.
X-Pool~\cite{gorti2022xpool} introduced a text-conditioned pooling module that aggregates frame features via cross-attention guided by textual context, while subsequent works extended this to hierarchical interaction~\cite{fang2023uatvr,patrick2020support,lee2020parameter} and multimodal cues such as audio~\cite{jeong2025learning}.
Despite such progress, the modality gap between compact textual descriptions and complex, dynamic video content remains an open challenge.

\subsection{Diffusion-based Retrieval}
Diffusion models~\cite{ho2020denoising,song2020score,dhariwal2021diffusion}
progressively refine random noise into structured data through a learned denoising process.
Although originally developed for high-fidelity generation,
diffusion mechanisms have been adopted for multimodal synthesis,
including text-to-image~\cite{ramesh2022hierarchical,saharia2022photorealistic}
and text-to-video~\cite{ho2022video,singer2022make}.
However, retrieval differs fundamentally from generation: it requires
stable similarity estimation rather than stochastic sampling.
DiffusionRet~\cite{jin2023diffusionret} attempted to model similarity
as a generative denoising process, but its random sampling introduces unnecessary variance.
DITS~\cite{li2023dits} improved upon this by employing a truncated, deterministic refinement
that begins from meaningful text embeddings instead of random noise,
demonstrating that diffusion-inspired iterative refinement
can enhance alignment without full generative sampling.
Nevertheless, DITS still operates on deterministic points--it refines feature vectors,
not distributions--and thus cannot explicitly capture uncertainty or diversity.

\subsection{Probabilistic and Uncertainty-Aware Retrieval}
The inherent one-to-many nature of multimodal alignment makes uncertainty modeling crucial.
Probabilistic embedding frameworks~\cite{chun2021probabilistic,chun2023improved,fang2023uatvr}
represent each modality as a Gaussian distribution whose mean encodes semantics
and variance reflects ambiguity.
UATVR~\cite{fang2023uatvr} extends this idea to TVR,
allowing variance to capture semantic diversity.
However, UATVR relies on sampled prototypes to compute similarity, which
partially collapses probabilistic representations into deterministic comparisons.
UAN~\cite{hao2023uncertainty} addresses uncertainty under cross-domain adaptation, but aligns features for
domain transfer rather than learning modality-intrinsic distributional refinement.
Recent fine-grained TVR methods such as Video-ColBERT~\cite{reddy2025video} and Hybrid-Tower~\cite{lan2025hybrid}
strengthen late interaction or pseudo-query interaction at the feature level.
DAB is complementary: it eliminates sampling and learns a deterministic bridge
over \((\mu,\log\sigma^2)\), directly refining both mean and variance with
directional KL costs.

\section{Methods}
\begin{figure*}[!t]
  \centering
  \includegraphics[width=\linewidth]{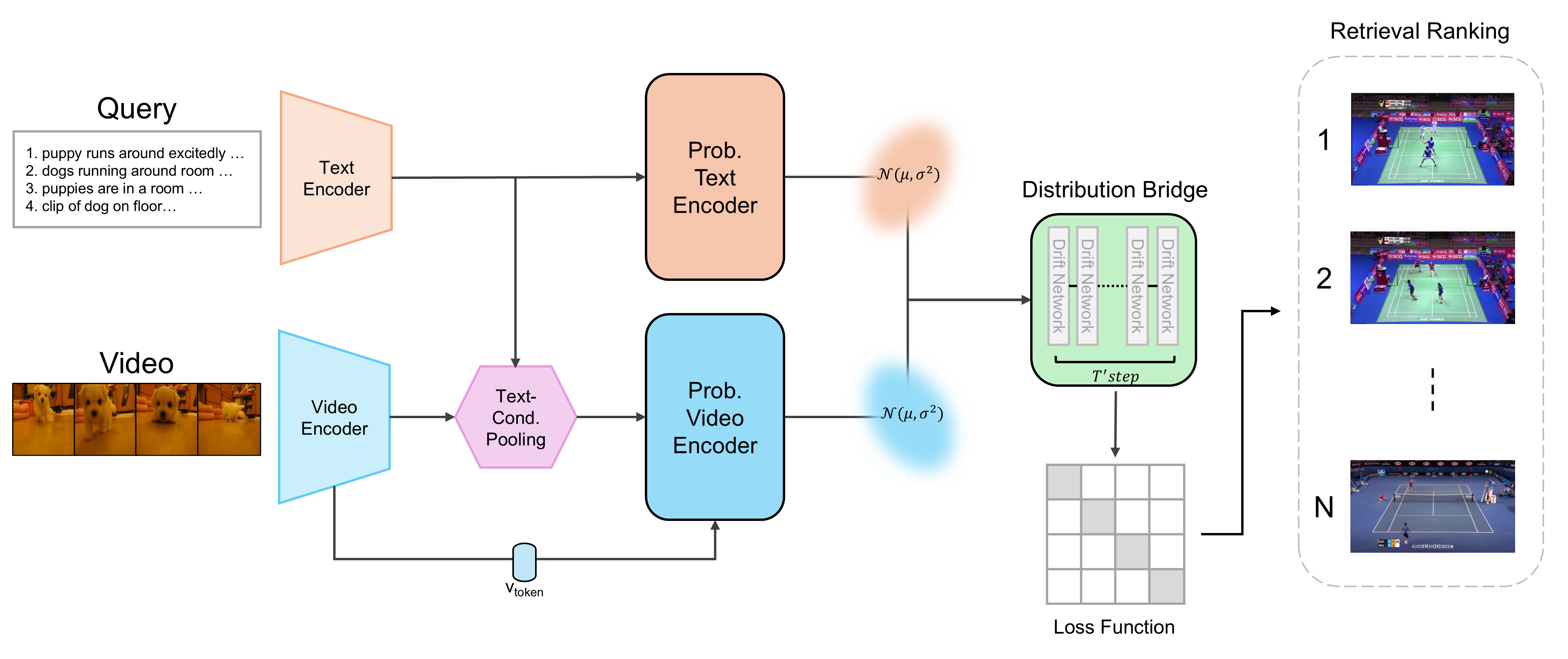}

  \caption{
    \textbf{Overall architecture of the proposed Distribution-Alignment Bridge (DAB).}
    DAB reformulates text-to-video retrieval as a distribution alignment framework.
    It encodes text and video embeddings via CLIP, transforms them into Gaussian distributions through a probabilistic encoder,
    and employs a deterministic distribution bridge to progressively align text distributions toward video distributions for contrastive training.
  }
  \label{fig:dab_overview}
\end{figure*}

\begin{figure*}[!t]
  \centering
  \includegraphics[width=\linewidth]{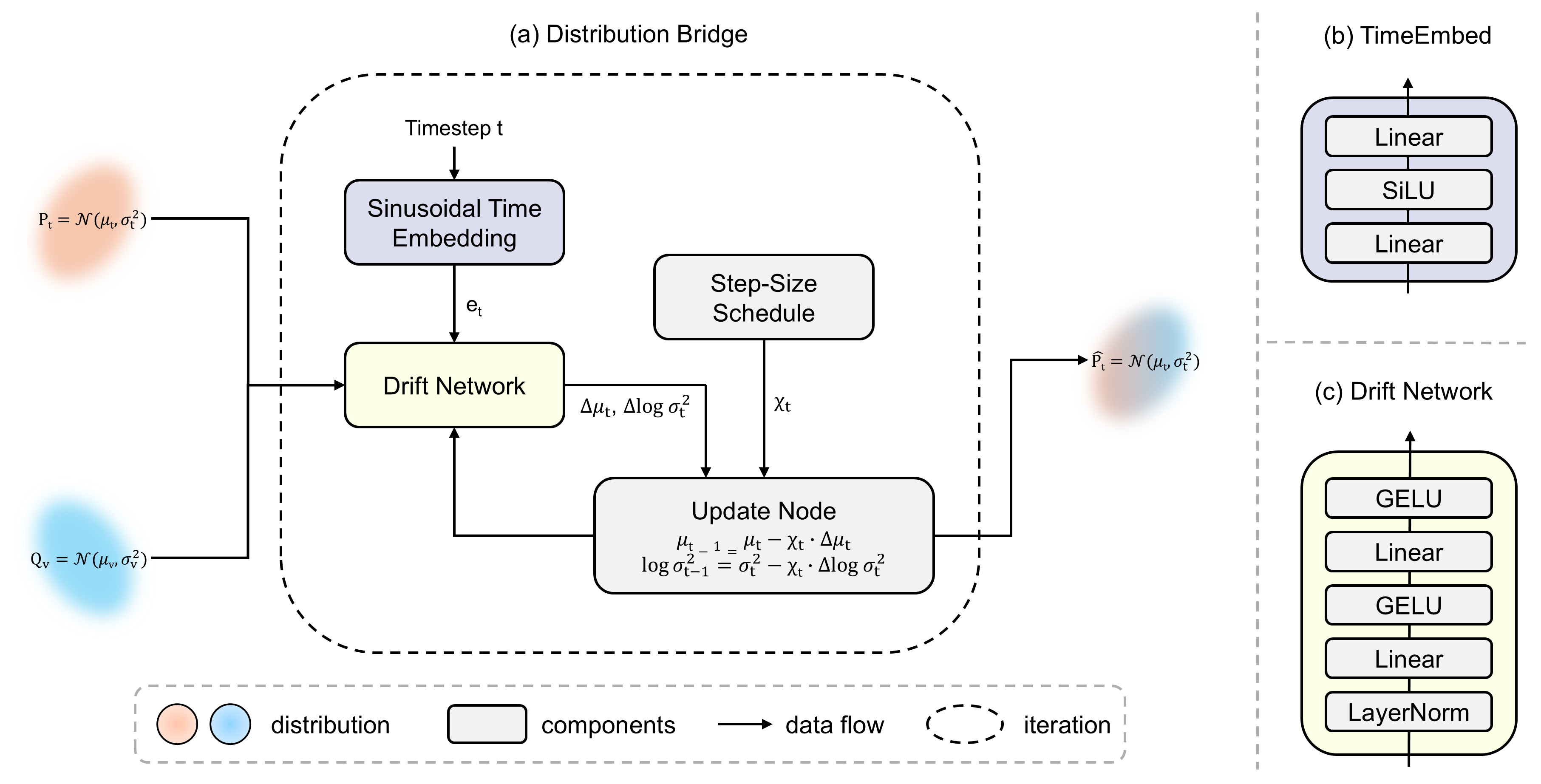}
  \caption{
    \textbf{Detailed structure of the Distribution Bridge module.}
    The bridge iteratively updates the mean and variance of text distributions toward those of the corresponding video distributions using a time-conditioned drift network.
    Through multi-step deterministic refinement, it achieves stable distribution alignment without stochastic sampling.
  }
  \label{fig:dab_bridge}
\end{figure*}

Our proposed DAB reframes text-to-video retrieval
from a deterministic point-matching problem into a distribution-alignment framework.
As illustrated in Fig.~\ref{fig:dab_overview}, the model consists of three main components:
(1) a foundational feature representation module that extracts text-conditioned video embeddings,
(2) a probabilistic encoder that converts embeddings into Gaussian distributions to explicitly model semantic uncertainty,
and (3) a distribution bridge that deterministically transforms the text distribution
toward the video distribution through iterative, drift-based refinement (see Fig.~\ref{fig:dab_bridge}).
This bridge serves as the core of our framework, enabling stable and efficient distribution alignment
without stochastic sampling.

\subsection{Foundational Feature Representation}
\subsubsection{CLIP Encoding.}
Given a batch of text--video pairs, the pretrained CLIP model
(ViT-B/32 and ViT-B/16) encodes text and video frames into a shared semantic space.
The text encoder \(G_\phi(\cdot)\) produces a sentence-level embedding
\(t \in \mathbb{R}^{B\times D}\),
while the visual encoder \(F_\theta(\cdot)\) extracts frame-level features
\(\{v_f\}_{f=1}^{F}\) for each video, resulting in
\(V \in \mathbb{R}^{B\times F\times D}\).
Here, \(B\) is the batch size, \(F\) is the number of sampled frames per video, and \(D\) is the embedding dimension.
These features serve as the foundation for subsequent distributional modeling.

\subsubsection{Text-Conditioned Frame Pooling.}
To obtain a representative video embedding that reflects textual context,
we apply a transformer-based cross-attention pooling module~\cite{gorti2022xpool}.
For each text--video pair, the text embedding \(t\) serves as the \textit{Query},
while frame embeddings \(\{v_f\}_{f=1}^{F}\) serves as the \textit{Keys} and \textit{Values}.
The resulting attention outputs are aggregated into
text-conditioned video representations \(v_c \in \mathbb{R}^{B\times B\times D}\),
where each element \(v_c^{(i,j)}\) corresponds to video \(j\) pooled conditioned on text \(i\).
Diagonal entries \(v_c^{(i,i)}\) thus represent the integrated embeddings of matched text--video pairs.
This cross-modal attention mechanism allows the model to focus on semantically relevant frames,
laying the foundation for the subsequent probabilistic encoding and alignment stages.

\begin{figure}[!t]
  \centering
  \includegraphics[width=0.5\linewidth]{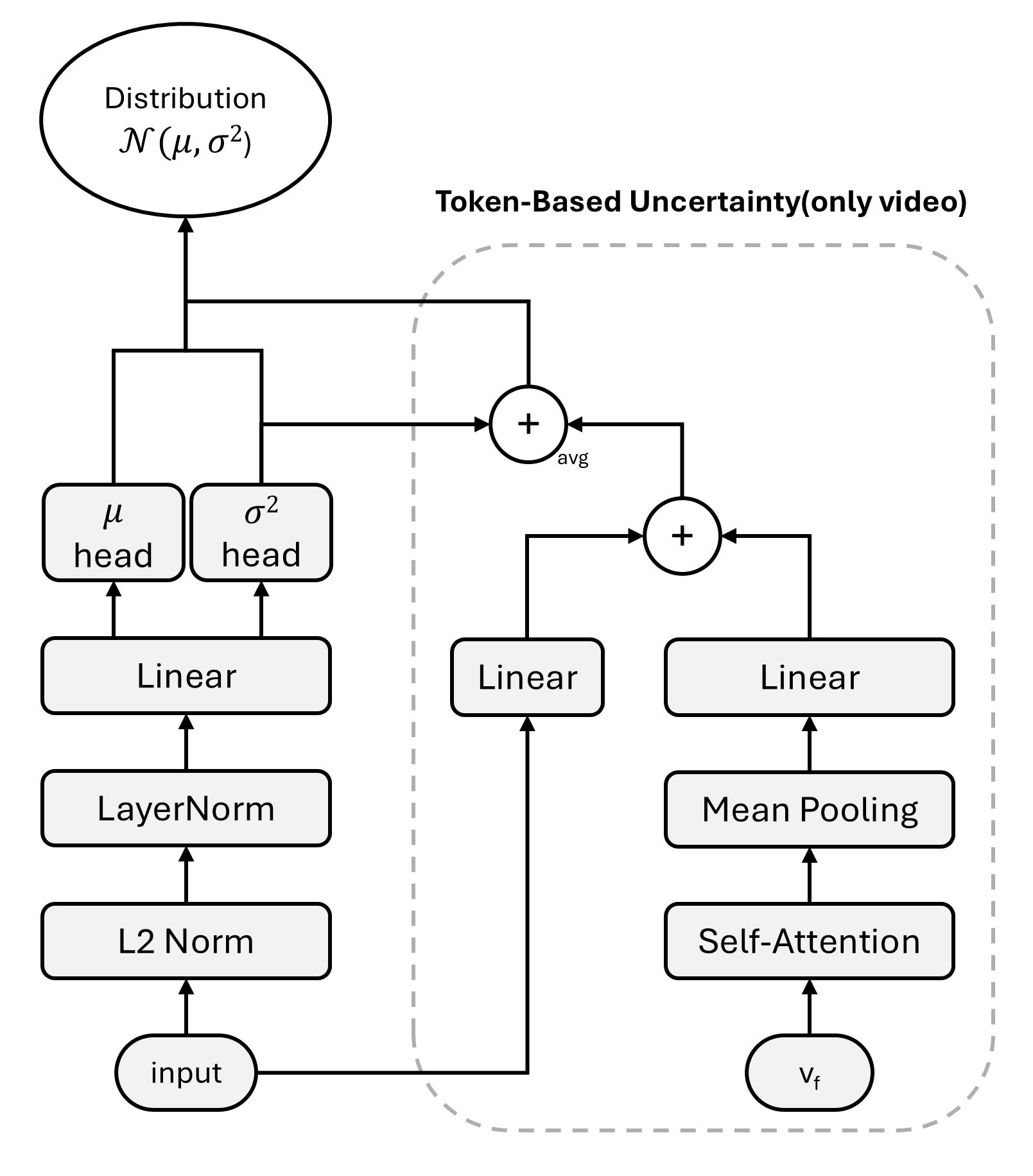}
  \caption{
    \textbf{Probabilistic Embedding Module.} An input embedding is normalized and linearly projected, then split into two heads that predict the mean (\(\mu\)) and variance (\(\sigma^{2}\)) of a Gaussian distribution. For video inputs, a token-based uncertainty branch refines the variance by applying self-attention and mean pooling over frame tokens. The fused variance and predicted mean form the final distribution \(\mathcal{N}(\mu,\sigma^{2})\).
  }

  \label{fig:prob_encoder}
\end{figure}

\subsection{Probabilistic Embedding Module}
\subsubsection{Probabilistic Head.}
As illustrated in Fig.~\ref{fig:prob_encoder}, the probabilistic embedding module predicts Gaussian parameters through separate mean and log-variance heads, with an additional token-based uncertainty branch for video inputs.
To capture semantic ambiguity, each text and video embedding is modeled
as a Gaussian distribution \(\mathcal{N}(\mu, \operatorname{diag}(\sigma^2))\)
rather than a deterministic point.
Before projection, the input embedding \(x\) is L2-normalized
as \(\tilde{x} = x / (\|x\|_2 + 10^{-6})\).
Two independent heads---each composed of a LayerNorm followed by a Linear layer---map
the normalized feature to the mean and log-variance:
\begin{equation}
  \begin{aligned}
    \mu          & = W_\mu\,\mathrm{LN}(\tilde{x}) + b_\mu,           \\
    \log\sigma^2 & = W_{\ell v}\,\mathrm{LN}(\tilde{x}) + b_{\ell v}.
  \end{aligned}
\end{equation}
The log-variance bias is initialized to \(-2.0\) to encourage low initial uncertainty,
and the predicted \(\log\sigma^2\) is numerically clipped within \([-6, 2]\) for stability.
The Probabilistic Head outputs \(\mu, \log\sigma^2 \in \mathbb{R}^{B \times D}\) for each modality.

\subsubsection{Token-Based Uncertainty for Video.}
When token-level representations \(\{v_f\}_{f=1}^{L}\) (e.g., frame embeddings) are available,
we introduce a \textit{Token-Based Uncertainty Head} to refine variance estimation based on intra-video variability.
The text-conditioned frame sequence is first processed by a single-head self-attention layer to capture contextual relations:
\begin{equation}
  \mathbf{H} = \mathrm{SelfAttn}(\{v_f\}),
  \quad \mathbf{H} \in \mathbb{R}^{B\times L\times D}.
\end{equation}

The token outputs are mean-pooled to summarize the spatio-temporal variability:
\begin{equation}
  \bar{h} = \frac{1}{L}\sum_{f=1}^{L} H_f,
  \quad \bar{h} \in \mathbb{R}^{B\times D}.
\end{equation}

Two linear projections are applied to the pooled feature and the global video embedding \(v_c\),
and their residual sum predicts the token-based log-variance:
\begin{equation}
  \log\sigma_{v,\text{tok}}^2 = W_1\bar{h} + b_1 + W_2v_c + b_2.
\end{equation}

Finally, the token-based and base variances are averaged to produce the final video uncertainty:
\begin{equation}
  \log\sigma_v^2 =
  \tfrac{1}{2}\big(\log\sigma_{v,\text{base}}^2 + \log\sigma_{v,\text{tok}}^2\big).
\end{equation}
For stability, all log-variance outputs are clipped to the range \([-6, 2]\) during training.

This formulation decouples semantic content (mean) from spatio-temporal uncertainty (variance),
allowing the model to represent diverse visual evidence across frames.
Both modalities share the same \textit{Probabilistic Encoder} that converts embeddings into Gaussian distributions.
For text, \((\mu_t, \log\sigma_t^2)\) is obtained from the sentence-level embedding,
whereas for video, \((\mu_v, \log\sigma_v^2)\) is derived from the pooled representation \(v_{\text{repr}}\)
and refined using frame-level tokens to capture temporal uncertainty.

\subsection{Distribution Bridge}

\subsubsection{Architecture.}
Given a text distribution \(P_t = \mathcal{N}(\mu_t, \sigma_t^2)\)
and a target video distribution \(Q_v = \mathcal{N}(\mu_v, \sigma_v^2)\),
the distribution bridge deterministically transforms \(P_t\) toward \(Q_v\)
through iterative refinement inspired by diffusion processes.
It uses discrete timesteps and updates parameters across \(T'\) steps.
The bridge comprises three components: a sinusoidal time embedding,
a drift network that predicts parameter offsets, and
a step-size schedule \(\chi_t\) controlling the update magnitude.

\subsubsection{Sinusoidal Time Embedding.}
Each timestep \(t\) is encoded as a continuous embedding \(e_t\)
via sinusoidal positional encoding followed by a lightweight MLP with SiLU activation:
\begin{equation}
  \begin{aligned}
    \mathrm{TimeEmbed}(t)
    = \mathrm{MLP}\!\left(
    \begin{bmatrix}
      \sin\!\left(\frac{t}{10000^{2k/d}}\right) \\
      \cos\!\left(\frac{t}{10000^{2k/d}}\right)
    \end{bmatrix}_{k=0}^{d/2-1}
    \right),
    \\
    e_t \in \mathbb{R}^{d_t}.
  \end{aligned}
\end{equation}
This embedding provides the drift network with temporal context,
allowing parameter updates to vary smoothly with the refinement stage. The time embedding is parameterized by \(d\), the embedding dimensionality, while \(k\) represents the index of the frequency components, ensuring that each dimension of \(e_t\) corresponds to a unique sinusoidal wavelength.

\subsubsection{Drift Network.}
The drift module (\textit{DriftDistMLP}) predicts mean and variance shifts
given the current text and target video distributions, together with the time embedding:
\begin{equation}
  \begin{aligned}
    [\Delta\mu_t,\, \Delta(\log\sigma_t^2)]
    = f_{\theta}^{\text{drift}}\!\big(
    [\mu_t,\, \sigma_t,\, \mu_v,\, \sigma_v,\, e_t]
    \big),
    \\
    \Delta\mu_t,\, \Delta(\log\sigma_t^2) \in \mathbb{R}^{B\times D}.
  \end{aligned}
\end{equation}
The network consists of a LayerNorm followed by two GELU-activated Linear layers,
and two independent output heads for \(\Delta\mu_t\) and \(\Delta(\log\sigma_t^2)\).
All output weights are zero-initialized to suppress abrupt drift during the early stages of training.

\subsubsection{Step-Size Schedule.}
Each refinement step is scaled by a precomputed coefficient \(\chi_t\)
derived from a linear \(\beta\)-schedule, following the truncated diffusion strategy~\cite{li2023dits}:
\begin{equation}
  \begin{aligned}
    \chi_t  & = \mathrm{Norm}\!\left(\sqrt{1 - \prod_{s=1}^{t} (1 - \beta_s)}\right), \\
    \beta_s & = \beta_{\min} + (\beta_{\max} - \beta_{\min})\tfrac{s}{T}.
  \end{aligned}
\end{equation}
Here \(\beta_{\min}=10^{-4}\) and \(\beta_{\max}=2\times10^{-2}\).
The resulting \(\chi_t\) values are normalized into the range \([0.2, 0.6]\)
for stable gradient scaling, and only the last \(T'\) truncated steps are used
to balance efficiency and stability.

\subsubsection{Iterative Refinement.}
At each timestep, the parameters are updated as:
\begin{equation}
  \begin{aligned}
    \mu_{t-1}          & = \mu_t - \chi_t \cdot \Delta\mu_t,                     \\
    \log\sigma_{t-1}^2 & = \log\sigma_t^2 - \chi_t \cdot \Delta(\log\sigma_t^2).
  \end{aligned}
\end{equation}
After each update, \(\log\sigma^2\) values are clipped to a fixed range \([-6,2]\)
to maintain numerical stability.
No stochastic noise is injected, producing a smooth and deterministic trajectory in distribution space.
After \(T'\) iterations, the refined distribution
\(\hat{P}_t = \mathcal{N}(\mu_0, \sigma_0^2)\)
is closely aligned with the target \(Q_v\).
This stepwise process enables gradual correction even for large initial modality gaps
and yields stable, sampling-free distribution alignment.

This update also provides a simple stability intuition. If the PAC-trained drift
approximates the residual \(x_t-x^\ast\), where \(x=(\mu,\log\sigma^2)\), then
\(x_{t-1}-x^\ast \approx (1-\chi_t)(x_t-x^\ast)\); since \(\chi_t\in[0.2,0.6]\),
each step behaves as a contraction with rate at most 0.8 under this approximation.
Together with zero-initialized drift heads and clipped log-variance, this keeps the
deterministic refinement trajectory bounded in practice.

\subsection{Optimization and Loss Function}
\subsubsection{Distributional Cost Matrix.}
Given a batch of \(B\) paired samples \((t_i, v_i)\),
we compute a pairwise divergence-based cost matrix between
the bridged text distributions \(\hat{P}_{t_i}\)
and the target video distributions \(Q_{v_j}\):
\begin{equation}
  C_{ij} = D\!\big(Q_{v_j} \,\|\, \hat{P}_{t_i}\big).
  \label{eq:cost_matrix}
\end{equation}
where \(D(\cdot\|\cdot)\) denotes a probabilistic divergence such as
the Kullback--Leibler (KL) divergence or, alternatively, a Wasserstein-based metric.

\subsubsection{Distribution Alignment Metric.}
For two diagonal Gaussian distributions
\(P=\mathcal{N}(\mu_P,\mathrm{diag}(\sigma_P^2))\)
and \(Q=\mathcal{N}(\mu_Q,\mathrm{diag}(\sigma_Q^2))\) in \(\mathbb{R}^D\),
the closed-form KL divergence is:
\begin{equation}
  D_{\mathrm{KL}}(P \parallel Q)
  = \tfrac{1}{2}\sum_{d=1}^{D}
  \left[
    \frac{(\mu_{Q,d}-\mu_{P,d})^2}{\sigma_{Q,d}^2}
    + \frac{\sigma_{P,d}^2}{\sigma_{Q,d}^2}
    - 1
    + \log\frac{\sigma_{Q,d}^2}{\sigma_{P,d}^2}
    \right].
  \label{eq:kl_closed_form}
\end{equation}
This asymmetric form \(D(Q_v\|\hat{P}_t)\) reflects the bridge direction
from text to video distributions, measuring how well the refined text distribution
approximates the target video distribution.
In practice, all log-variance outputs are clipped within a fixed range
for numerical stability.
The KL divergence effectively penalizes both mean and variance mismatches,
providing stable and uncertainty-aware updates for deterministic bridge refinement.

\subsubsection{Probabilistic Alignment Contrastive (PAC) Loss.}
Using the cost matrix \(C\), we adopt a bidirectional contrastive objective over
directional KL costs:
\begin{equation}
  \mathcal{L}_{\mathrm{PAC}}
  = -\frac{1}{B}\sum_{i=1}^{B}
  \left[
    \log \frac{\exp(-C_{ii}/\tau)}{\sum_{j=1}^{B}\exp(-C_{ij}/\tau)}
    +
    \log \frac{\exp(-C_{ii}/\tau)}{\sum_{j=1}^{B}\exp(-C_{ji}/\tau)}
    \right].
  \label{eq:pac}
\end{equation}
where \(\tau\) is a temperature parameter controlling contrastive sharpness.
The first term ranks videos for each text, and the second ranks texts for each
video. Importantly, the pairwise cost remains directional,
\(C_{ij}=D_{\mathrm{KL}}(Q_{v_j}\|\hat{P}_{t_i})\); thus DAB does not optimize
symmetric KL or JS divergence. The bidirectional objective only requires matched
directional costs to be smaller than in-batch negatives, and does not force text
and video variances to become identical.

\section{Experiments}
\subsection{Experimental Settings}
\subsubsection{Datasets and Metrics.}
We evaluate DAB on three public benchmarks:
MSR-VTT, MSVD, and VATEX.
(1) MSR-VTT~\cite{xu2016msrvtt} contains 10K videos with 20 captions each.
We train on the \textit{9K} split and report on the 1K-A test set~\cite{luo2022clip4clip}.
(2) MSVD~\cite{chen2011collecting} has 1{,}970 YouTube videos with \(\sim\)80K captions
using the standard \(1{,}200/100/670\) split.
(3) VATEX~\cite{wang2019vatex} includes \(\sim\)35K videos with multiple captions;
we follow the split in~\cite{chen2020fine}.
We report \textit{Recall@\(\{1,5,10\}\)}, \textit{MdR}, and \textit{MnR} (higher recall and lower ranks are better).

\begin{table*}[]
  \caption{Text-to-Video Retrieval on \textbf{MSR-VTT}.}
  \centering
  \resizebox{\linewidth}{!}{
    \begin{tabular}{@{}lccccccc@{}}
      \toprule
      Method                                  & Venue             & R@1           & R@5           & R@10          & MdR        & MnR           & RSum           \\
      \midrule
      \textit{\textbf{CLIP-ViT-B/32}}         &                   &               &               &               &            &               &                \\
      CLIP4Clip~\cite{luo2022clip4clip}       & Neurocomputing'22 & 44.5          & 71.4          & 81.6          & 2          & 15.3          & 197.5          \\
      X-Pool~\cite{gorti2022xpool}            & CVPR'22           & 46.9          & 72.8          & 82.2          & 2          & 14.3          & 201.9          \\
      UATVR~\cite{fang2023uatvr}              & ICCV'23           & 47.5          & 73.9          & 83.5          & 2          & 12.9          & 204.9          \\
      Video-ColBERT~\cite{reddy2025video}     & CVPR'25           & 48.1          & 74.9          & 83.9          & --         & --            & 206.9          \\
      PIG~\cite{lan2025hybrid}                & ICCV'25           & 48.6          & 72.8          & 81.6          & --         & --            & 203.0          \\
      DiffusionRet~\cite{jin2023diffusionret} & ICCV'23           & 49.0          & 75.2          & 82.7          & 2          & 12.1          & 206.9          \\
      UCOFIA~\cite{wang2023unified}           & ICCV'23           & 49.4          & 72.1          & --            & --         & 12.9          & --             \\
      T-MASS~\cite{wang2024text}              & CVPR'24           & 50.2          & 75.3          & 85.1          & 2          & 11.9          & 210.6          \\
      AVIGATE~\cite{jeong2025learning}        & CVPR'25           & 50.2          & 74.3          & 83.2          & --         & --            & 207.7          \\
      NarVid~\cite{hur2025narrating}          & CVPR'25           & 50.8          & 76.4          & 84.7          & 1          & 11.6          & 211.9          \\
      DITS~\cite{li2023dits}                  & NeurIPS'24        & 51.9          & 75.7          & 84.6          & 1          & 11.6          & 212.2          \\

      \midrule
      \textbf{DAB}                            & Proposed          & \textbf{56.2} & \textbf{85.4} & \textbf{92.4} & \textbf{1} & \textbf{4.1}  & \textbf{234.0} \\

      \midrule
      \textit{\textbf{CLIP-ViT-B/16}}         &                   &               &               &               &            &               &                \\
      X-Pool~\cite{gorti2022xpool}            & CVPR'22           & 48.2          & 73.7          & 82.6          & 2          & 12.7          & 204.5          \\
      UATVR~\cite{fang2023uatvr}              & ICCV'23           & 50.8          & 76.3          & 85.5          & 1          & 12.4          & 212.6          \\
      Video-ColBERT~\cite{reddy2025video}     & CVPR'25           & 51.0          & 77.1          & 85.5          & --         & --            & 213.6          \\
      AVIGATE~\cite{jeong2025learning}        & CVPR'25           & 52.1          & 76.4          & 85.2          & --         & --            & 213.7          \\
      NarVid~\cite{hur2025narrating}          & CVPR'25           & 52.7          & 77.7          & 85.6          & 1          & 12.3          & 216.0          \\
      T-MASS~\cite{wang2024text}              & CVPR'24           & 52.7          & 77.1          & 85.6          & 1          & 10.5          & 215.4          \\
      DITS~\cite{li2023dits}                  & NeurIPS'24        & 55.0          & 79.8          & 87.1          & 1          & 10.0          & 221.9          \\
      \midrule
      \textbf{DAB}                            & Proposed          & \textbf{57.6} & \textbf{86.2} & \textbf{92.4} & \textbf{1} & \textbf{4.17} & \textbf{236.2} \\

      \bottomrule
    \end{tabular}
  }
  \label{tab:msrvtt_main}
\end{table*}

\begin{table*}[]
  \caption{Text-to-Video Retrieval on \textbf{MSVD}.
    \textcolor{gray}{Gray row} denotes ViT-B/16 backbone.}
  \centering
  \resizebox{\linewidth}{!}{
    \begin{tabular}{@{}lccccccc@{}}
      \toprule
      Method                                           & Venue                     & R@1                    & R@5                    & R@10                   & MdR                 & MnR                  & RSum                    \\
      \midrule
      CLIP4Clip \cite{luo2022clip4clip}                & Neurocomputing'22         & 45.2                   & 75.5                   & 84.3                   & 2                   & 10.0                 & 205.0                   \\
      Video-ColBERT \cite{reddy2025video}              & CVPR'25                   & 46.0                   & 75.0                   & 84.0                   & --                  & --                   & 205.0                   \\
      UATVR \cite{fang2023uatvr}                       & ICCV'23                   & 46.0                   & 76.3                   & 85.1                   & 2                   & --                   & 207.4                   \\
      X-Pool \cite{gorti2022xpool}                     & CVPR'22                   & 47.2                   & 77.4                   & 86.0                   & 2                   & 9.3                  & 210.6                   \\
      UCOFIA \cite{wang2023unified}                    & ICCV'23                   & 47.4                   & 77.6                   & --                     & --                  & --                   & --                      \\
      DiffusionRet \cite{jin2023diffusionret}          & ICCV'23                   & 47.9                   & 77.2                   & 84.8                   & 2                   & 15.6                 & 209.9                   \\
      PIG \cite{lan2025hybrid}                         & ICCV'25                   & 47.9                   & 75.9                   & 83.4                   & --                  & --                   & 207.2                   \\
      Cap4Video \cite{wu2023cap4video}                 & CVPR'23                   & 51.8                   & 80.8                   & 88.3                   & 1                   & --                   & 220.9                   \\
      \textcolor{gray}{NarVid \cite{hur2025narrating}} & \textcolor{gray}{CVPR'25} & \textcolor{gray}{53.1} & \textcolor{gray}{81.4} & \textcolor{gray}{88.8} & \textcolor{gray}{1} & \textcolor{gray}{--} & \textcolor{gray}{223.3} \\
      \midrule
      \textbf{DAB (ViT-B/32)}                          & Proposed                  & \textbf{54.5}          & \textbf{87.8}          & \textbf{93.9}          & \textbf{1}          & \textbf{3.4}         & \textbf{236.2}          \\
      \bottomrule
    \end{tabular}
  }

  \label{tab:msvd_main}
\end{table*}

\begin{table}[]
  \caption{Text-to-Video Retrieval on \textbf{VATEX}.
    \textcolor{gray}{Gray rows} denote ViT-B/16 backbone.}
  \centering
  \resizebox{\linewidth}{!}{
    \begin{tabular}{@{}lccccccc@{}}
      \toprule
      Method                                                & Venue                     & R@1                    & R@5                    & R@10                   & MdR                 & MnR                  & RSum                    \\
      \midrule
      CLIP4Clip \cite{luo2022clip4clip}                     & Neurocomputing'22         & 55.9                   & 89.2                   & 95.0                   & 1                   & --                   & 240.1                   \\
      X-Pool \cite{gorti2022xpool}                          & CVPR'22                   & 60.0                   & 90.0                   & 95.0                   & 1                   & 3.8                  & 245.0                   \\
      UCOFIA \cite{wang2023unified}                         & ICCV'23                   & 61.1                   & 90.5                   & --                     & 1                   & 3.4                  & --                      \\
      UATVR \cite{fang2023uatvr}                            & ICCV'23                   & 61.3                   & 91.0                   & 95.6                   & 1                   & 3.3                  & 247.9                   \\
      T-MASS \cite{wang2024text}                            & CVPR'24                   & 63.0                   & 92.3                   & 96.4                   & 1                   & 3.2                  & 251.7                   \\
      AVIGATE \cite{jeong2025learning}                      & CVPR'25                   & 63.1                   & 90.7                   & 95.5                   & 1                   & --                   & 249.3                   \\
      PIG \cite{lan2025hybrid}                              & ICCV'25                   & 64.0                   & 91.5                   & 96.6                   & 1                   & --                   & 252.1                   \\
      DITS \cite{li2023dits}                                & NeurIPS'24                & 64.1                   & 92.7                   & 97.0                   & 1                   & 2.9                  & 253.8                   \\
      \textcolor{gray}{Video-ColBERT \cite{reddy2025video}} & \textcolor{gray}{CVPR'25} & \textcolor{gray}{66.8} & \textcolor{gray}{92.9} & \textcolor{gray}{96.8} & \textcolor{gray}{1} & \textcolor{gray}{--} & \textcolor{gray}{256.5} \\
      \textcolor{gray}{NarVid \cite{hur2025narrating}}      & \textcolor{gray}{CVPR'25} & \textcolor{gray}{68.4} & \textcolor{gray}{94.0} & \textcolor{gray}{97.1} & \textcolor{gray}{1} & \textcolor{gray}{--} & \textcolor{gray}{259.5} \\
      \midrule
      \textbf{DAB (ViT-B/32)}                               & Proposed                  & \textbf{65.4}          & \textbf{93.1}          & \textbf{97.2}          & \textbf{1}          & \textbf{2.4}         & \textbf{255.7}          \\
      \textbf{DAB (ViT-B/16)}                               & Proposed                  & \textbf{70.7}          & \textbf{95.7}          & \textbf{98.3}          & \textbf{1}          & \textbf{1.9}         & \textbf{264.7}          \\
      \bottomrule
    \end{tabular}
  }

  \label{tab:vatex_main}
\end{table}

\subsubsection{Implementation Details.}
We build on X-Pool~\cite{gorti2022xpool}.
Frames are resized to \(224{\times}224\) and we uniformly sample \(F{=}12\) frames per clip.
CLIP ViT-B/32 and CLIP ViT-B/16~\cite{radford2021learning} are used for both text and video encoders,
with a \(d{=}512\) projection.
A lightweight transformer pooling aggregates temporal information;
dropout is \(0.3\) for MSR-VTT/VATEX and \(0.4\) for MSVD.
We use AdamW~\cite{loshchilov2017decoupled} with weight decay \(\lambda{=}0.02\) and a warm-up ratio of \(0.1\).
Learning rates are \(1\mathrm{e}{-6}\) (CLIP) and \(1\mathrm{e}{-4}\) (probabilistic/bridge modules).
Batch size is \(B{=}32\); seed is 24.
We train for 10 epochs on all datasets.

For the bridge, we adopt \(N{=}1000\) total steps and truncated steps \(T'{=}32\) with a linear schedule
\(\beta_t \in [10^{-4},\,2\times 10^{-2}]\) following diffusion-inspired alignment~\cite{fang2023uatvr,li2023dits}.
Parameters \(\Theta=\{\theta,\phi,\gamma\}\) are optimized jointly.
Inference uses global text/video chunk sizes of 128/512.
All experiments are in PyTorch on a single NVIDIA A6000 GPU.

\subsection{Comparisons}
We compare DAB with recent SOTA models on MSR-VTT, MSVD, and VATEX,
including discriminative CLIP-based~\cite{luo2022clip4clip,gorti2022xpool},
probabilistic~\cite{fang2023uatvr,wang2024text},
and diffusion-inspired methods~\cite{jin2023diffusionret,li2023dits}.
Unless noted, results are obtained under consistent settings.

\subsubsection{MSR-VTT.}
As shown in Table~\ref{tab:msrvtt_main},
DAB achieves the best overall performance among all diffusion-based and probabilistic approaches.
It substantially improves R@1 to 56.2, surpassing DITS~\cite{li2023dits} by +4.3,
and reduces MnR from 11.6 to 4.14,
representing a 64\% relative reduction.
This indicates that DAB forms a globally coherent embedding space
in which relevant items cluster near the top ranks, rather than relying on isolated top-1 matches.
Compared to probabilistic UATVR~\cite{fang2023uatvr} and generative DiffusionRet~\cite{jin2023diffusionret},
which rely on stochastic sampling,
DAB's deterministic, sampling-free distribution alignment produces
both stable convergence and strong generalization.
Even against T-MASS~\cite{wang2024text}, which introduces stochastic text diffusion,
DAB achieves higher recall, confirming the efficacy of its distribution-to-distribution refinement.

\subsubsection{MSVD.}
Table~\ref{tab:msvd_main} shows consistent performance gains across all recall metrics.
DAB reaches 54.5/87.8/93.9 at R@1/5/10,
outperforming previous diffusion-based or probabilistic models by large margins.
Despite the smaller dataset scale and simpler sentence structures,
DAB's deterministic bridge improves R@1 over CLIP4Clip~\cite{luo2022clip4clip} by +9.3
and reduces MnR to 3.4.
We found that higher dropout (\(0.4\)) and slightly longer training (10 epochs)
help mitigate overfitting on MSVD,
while the coarse-to-fine refinement of DAB naturally regularizes the feature space.

\subsubsection{VATEX.}
As shown in Table~\ref{tab:vatex_main},
DAB further improves the R@1 score to 65.4,
surpassing DITS by +1.3 and T-MASS by +2.4,
while achieving the lowest MnR among B/32-style comparisons. With ViT-B/16, DAB reaches
70.7/95.7/98.3 at R@1/5/10 and obtains the best overall VATEX result.
The gains on VATEX suggest that DAB benefits from richer textual
descriptions. Attribute-rich captions provide more semantic cues for estimating
uncertainty, while the deterministic bridge tightens cross-modal alignment
without collapsing modality-specific variance. This helps explain the consistent
improvements in both recall and MnR.
These gains are achieved without post-processing techniques such as inverted softmax
or dual normalization, showing that sampling-free distribution alignment alone
is highly effective.

\medskip
In summary, DAB delivers strong state-of-the-art performance across MSR-VTT, MSVD, and VATEX, achieving leading results on MSR-VTT and MSVD among the compared methods and the best overall VATEX result with the ViT-B/16 backbone. Beyond higher recall, the pronounced reductions in MnR consistently show that DAB constructs a globally coherent and uncertainty-aware embedding space through distribution-level refinement.

\subsection{Ablation Study}
We analyze the effect of key components and hyperparameters of DAB on MSR-VTT (trained for 5 epochs) with CLIP ViT-B/32, reporting text-to-video results.

\begin{table*}[t]
  \caption{Accuracy-efficiency trade-off for truncated steps \(T'\) on MSR-VTT (5 epochs).}
  \centering
  \begin{tabular}{ccccccccc}
    \toprule
    \(T'\) & R@1           & R@5           & R@10          & MdR & MnR          & RSum           & Latency(s) & FLOPs(T) \\
    \midrule
    8      & 49.0          & 82.0          & 89.0          & 2   & 5.8          & 220.0          & 39.5       & 69.73    \\
    16     & 50.6          & 80.7          & 89.4          & 1   & 5.6          & 220.7          & 43.9       & 139.47   \\
    32     & \textbf{52.6} & 82.7          & 90.8          & 1   & \textbf{5.3} & 226.1          & 51.8       & 278.93   \\
    64     & 51.9          & \textbf{84.2} & \textbf{91.1} & 1   & 5.6          & \textbf{227.2} & 68.9       & 557.85   \\
    \bottomrule
  \end{tabular}
  \label{tab:ablation_t-steps}
\end{table*}

\subsubsection{Effect of Truncated Steps \(T'\).}
Table~\ref{tab:ablation_t-steps} examines the impact of the truncated refinement steps.
As \(T'\) increases, the iterative bridge refines alignment more precisely,
and performance steadily improves, peaking at \(T'=32\).
Although \(T'=64\) slightly improves R@5 and R@10, its latency increases from
51.8s to 68.9s compared with \(T'=32\). FLOPs scale nearly linearly with \(T'\)
because each step applies the same lightweight drift MLP.
We therefore adopt \(T'=32\) as the default setting, as it achieves the best R@1
while maintaining a favorable accuracy-efficiency trade-off.

\begin{wraptable}{r}{0.4\textwidth}
  \vspace{-35pt}
  \centering
  \small
  \caption{Component ablation and distributional cost comparison on MSR-VTT (5 epochs).}
  \label{tab:ablation_probabilistic}

  \begin{tabular*}{\linewidth}{@{}l@{\extracolsep{\fill}}cccc@{}}
    \multicolumn{5}{c}{\textbf{(a) Component Analysis}} \\
    \toprule
    Method    & R@1   & R@5   & R@10  & MnR                \\
    \midrule
    Baseline  & 46.9 & 72.8 & 82.2 & 14.3               \\
    w/ Prob.  & 44.3 & 76.6 & 87.2 & 7.4                \\
    w/ Bridge & 43.2 & 76.6 & 86.3 & 8.5                \\
    DAB       & 52.6 & 82.7 & 90.8 & 5.3                \\
    \bottomrule
  \end{tabular*}

  \vspace{6pt}

  \begin{tabular*}{\linewidth}{@{}l@{\extracolsep{\fill}}cccc@{}}
    \multicolumn{5}{c}{\textbf{(b) Distributional Cost}} \\
    \toprule
    Cost  & R@1   & R@5   & R@10  & MnR                     \\
    \midrule
    KL    & 52.6 & 82.7 & 90.8 & 5.3                     \\
    \(W_2\) & 46.8 & 79.7 & 88.4 & 6.6                     \\
    \bottomrule
  \end{tabular*}

  \vspace{-15pt}
\end{wraptable}

\subsubsection{Effect of Probabilistic Embedding and Distribution Bridge.}
The component ablation in Table~\ref{tab:ablation_probabilistic} highlights the contribution of each module.
Starting from X-Pool, incorporating the probabilistic embedding substantially reduces MnR
and enhances overall recall by modeling semantic uncertainty.
Applying the bridge alone produces limited improvement, suggesting that
distribution-level refinement is most effective only when guided by uncertainty.
Combining both components results in the best overall performance,
validating that DAB's strength lies in its deterministic yet uncertainty-aware
distribution refinement.

The cost ablation further shows that replacing directional KL with W2 degrades
R@1 from 52.6 to 46.8 and increases MnR from 5.3 to 6.6, supporting KL as a
more ranking-sensitive divergence for DAB.

\subsection{Uncertainty and Semantic-Neighborhood Diagnostics}

\begin{figure*}[t]
  \centering
  \includegraphics[width=\linewidth]{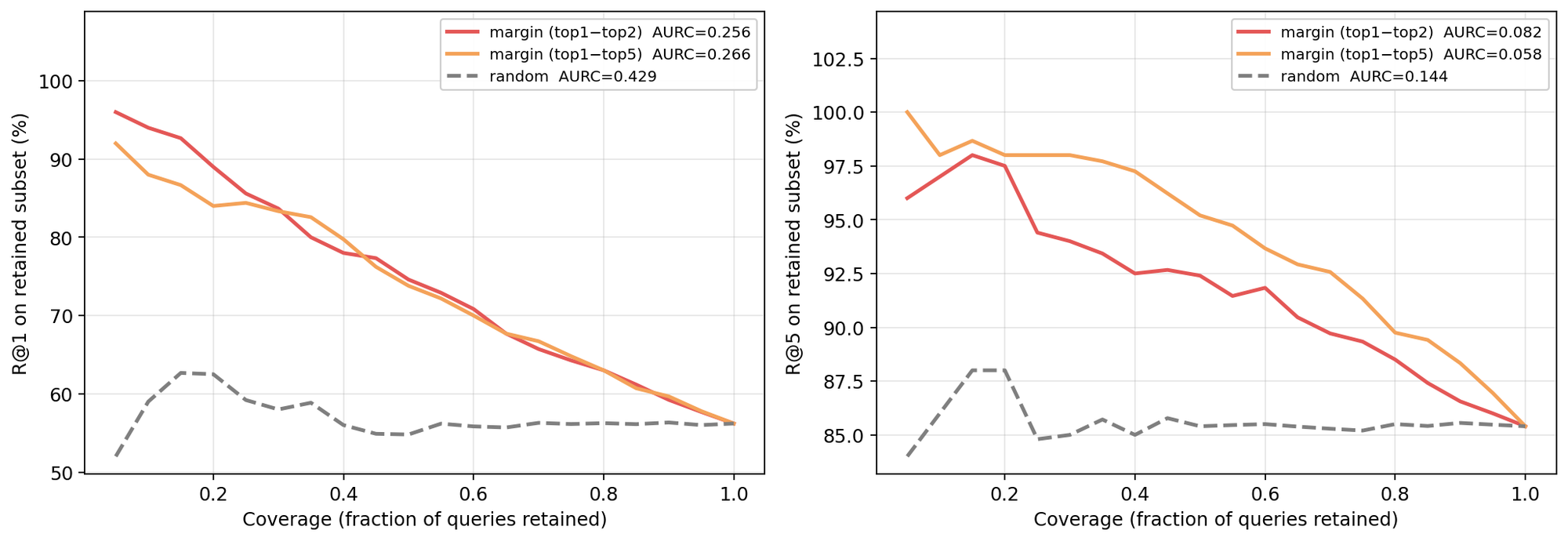}
  \caption{
    \textbf{Uncertainty calibration on MSR-VTT.} Queries are ranked by the
    bridge-induced KL margin between the best and competing retrieved
    candidates, and retrieval accuracy is evaluated under selective prediction.
    Larger margins correspond to more reliable retrieval decisions, yielding
    substantially lower risk and AURC than random ordering. This indicates
    that DAB provides a calibrated confidence signal through distributional
    refinement.
  }
  \label{fig:calibration}
\end{figure*}

\begin{table}[h]
  \centering
  \begin{minipage}[t][5.6cm][t]{0.48\linewidth}
    \centering
    \caption{\textbf{Learned variance asymmetry on MSR-VTT.} Avg. \(\sigma^2\)
      denotes the average variance of each distribution after training.
      The bridge variance remains between text and video variances,
      indicating that DAB preserves modality-specific uncertainty rather
      than collapsing both modalities to a shared variance scale during
      distribution refinement.}
    \label{tab:var_asym}

    \setlength{\tabcolsep}{3.5pt}
    \begin{tabular}{lccc}
      \toprule
                        & Text  & Bridge & Video \\
      \midrule
      Avg. \(\sigma^2\) & 0.507 & 0.935  & 2.979 \\
      Ratio             & 1.00  & 1.84   & 5.87  \\
      \bottomrule
    \end{tabular}

    \vfill
  \end{minipage}
  \hfill
  \begin{minipage}[t][5.6cm][t]{0.48\linewidth}
    \centering
    \caption{\textbf{Local semantic positives recovered in the top-10 on MSR-VTT.}
      A video is counted as a semantic positive if its visual feature has
      a cosine similarity of at least \(\tau\) with the ground-truth video.
      DAB consistently retrieves more semantic positives than random
      ranking across all thresholds tested.}
    \label{tab:semantic_neighborhood}

    \setlength{\tabcolsep}{3pt}
    \begin{tabular}{ccccc}
      \toprule
      \(\tau\) & Elig. & Pos.  & Rand. & DAB  \\
      \midrule
      0.3      & 921   & 50.05 & 0.50  & 5.06 \\
      0.4      & 864   & 17.75 & 0.18  & 3.22 \\
      0.5      & 655   & 7.28  & 0.07  & 2.18 \\
      \bottomrule
    \end{tabular}

    \vfill
  \end{minipage}

  \vspace{-15pt}
\end{table}

We further examine whether DAB's distributional variables provide meaningful
uncertainty signals beyond recall. First, the learned variances remain
modality-asymmetric at convergence. As shown in Table~\ref{tab:var_asym},
the average video variance is 5.87 times that of the text variance, while
the bridge variance lies between the two endpoints. This indicates that DAB does
not homogenize text and video uncertainty, but navigates between a low-entropy
textual anchor and a higher-entropy video distribution.

For uncertainty calibration, we use the bridge-induced KL margin as a confidence
signal. When queries are sorted by the top-1--top-2 KL margin, R@1 reaches
85.6\% at 25\% coverage, compared with 56.2\% under random ordering, and
further reaches 94.0\% at 10\% coverage. This shows that DAB's uncertainty
signal emerges from the bridge-induced distributional gap rather than from text
variance alone. As illustrated in Fig.~\ref{fig:calibration}, This trend is consistent across the full risk--coverage curve. Compared with
random ordering, KL-margin ranking substantially reduces the area under the
risk--coverage curve (AURC), from 0.429 to 0.256 for R@1 and from 0.144 to
0.082 for R@5, indicating that the bridge-induced margin provides a reliable
confidence estimate across a wide range of operating points.

Finally, we evaluate local one-to-many behavior by counting semantic positives
in the top-10, where positives are videos whose visual features have cosine
similarity at least \(\tau\) to the ground-truth video. As shown in
Table~\ref{tab:semantic_neighborhood}, DAB retrieves substantially more local
semantic positives than random ranking across all thresholds. This supports
local semantic-neighborhood organization rather than exhaustive global
multimodal coverage.

\begin{figure*}[h]
  \centering
  \includegraphics[width=\linewidth]{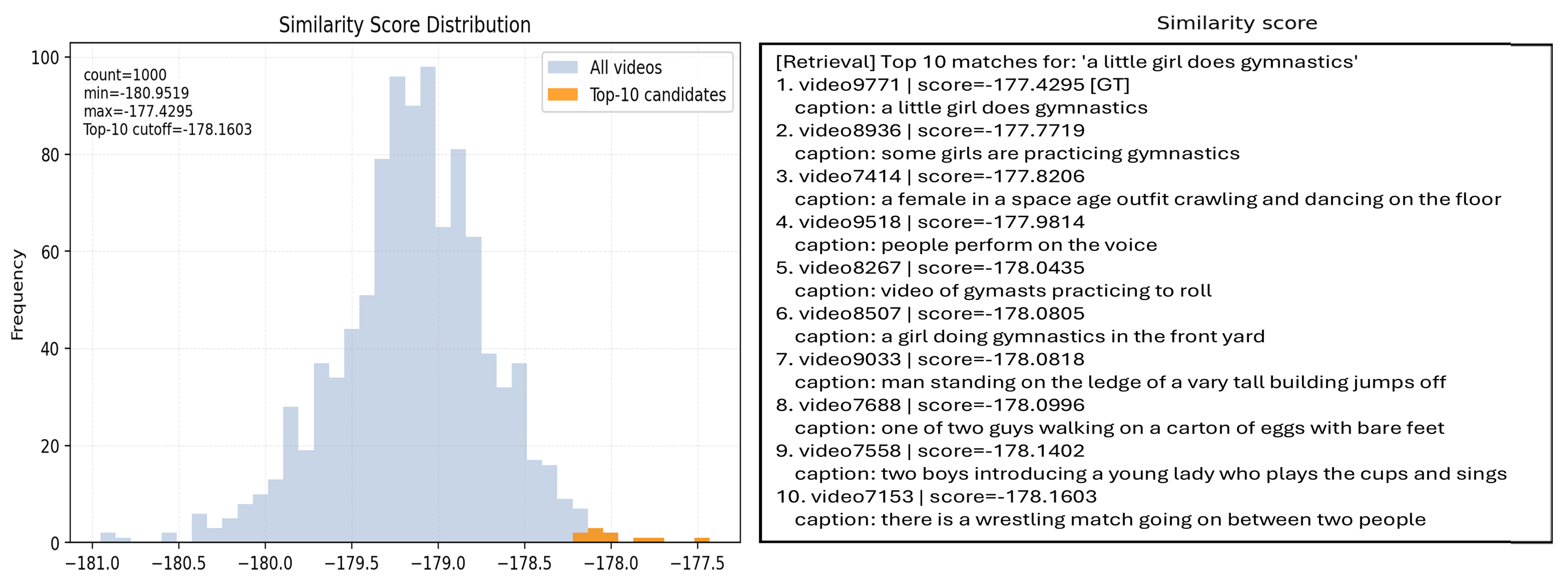}
  \caption{
    \textbf{Qualitative local-neighborhood example.}
    For the query ``\textit{a little girl does gymnastics}'', DAB ranks the annotated
    ground-truth video first and retrieves multiple gymnastics-related clips near
    the top. The highlighted top-ranked region provides an instance-level
    illustration of the local semantic-neighborhood behavior quantified in
    Table~\ref{tab:semantic_neighborhood}.
  }
  \label{fig:sim_distribution}
\end{figure*}

This aggregate trend is also visible at the instance level.
Fig.~\ref{fig:sim_distribution} visualizes the retrieval-score distribution for the
query ``\textit{a little girl does gymnastics}'' over 1,000 MSR-VTT test videos.
Scores are negative KL-based costs, so larger values indicate closer
distributional alignment. DAB ranks the annotated ground-truth video first and
also places several gymnastics-related clips among the top results, including
``\textit{some girls are practicing gymnastics}'',
``\textit{video of gymnasts practicing to roll}'', and
``\textit{a girl doing gymnastics in the front yard}''. This example qualitatively
illustrates the same local-neighborhood behavior measured in
Table~\ref{tab:semantic_neighborhood}: DAB does not only separate one annotated
positive from unrelated negatives, but organizes semantically related videos near
the query in the ranking space.

\section{Conclusion}
We introduced DAB, a framework that recasts text-to-video retrieval as
distribution-to-distribution alignment. By encoding both modalities as Gaussians
and refining them via a deterministic Distribution Bridge, DAB learns a
sampling-free trajectory in the space of \((\mu,\log\sigma^2)\), jointly updating
semantic centers and uncertainty scales. Central to this success is a bidirectional
contrastive objective over directional KL costs, which improves ranking-sensitive
alignment compared with W2 and preserves meaningful uncertainty signals throughout refinement. Empirically, DAB achieves strong results on MSR-VTT, MSVD, and VATEX. Further diagnostics show that learned variance
asymmetry is preserved, bridge-induced KL margins provide calibrated retrieval
confidence, and DAB organizes local semantic neighborhoods for ambiguous
queries. These results support DAB as an uncertainty-aware retrieval framework
whose uncertainty signal emerges from distributional refinement rather than from
variance magnitude alone.

\section*{Acknowledgements}
This work was supported by
the National Research Foundation of Korea(NRF)
grant funded by the Korea government(MSIT)(No.
RS-2025-02214941).
This research was supported by the Regional Innovation System \& Education(RISE) Glocal 30 program through the Daegu RISE Center, funded by the Ministry of Education(MOE) and the Daegu, Republic of Korea(2025-RISE-03-001).

%
%
\bibliographystyle{splncs04}
\bibliography{main}

@inproceedings{fang2023uatvr,
  author    = {Fang, Bo and Wu, Wenhao and Liu, Chang and Zhou, Yu and Song, Yuxin and Wang, Weiping and Shu, Xiangbo and Ji, Xiangyang and Wang, Jingdong},
  title     = {Uatvr: Uncertainty-adaptive text-video retrieval},
  booktitle = {Proceedings of the IEEE/CVF International Conference on Computer Vision},
  pages     = {13723-13733},
  year      = {2023},
  type      = {Conference Proceedings}
}

@inproceedings{gorti2022xpool,
  author    = {Gorti, Satya Krishna and Vouitsis, Noël and Ma, Junwei and Golestan, Keyvan and Volkovs, Maksims and Garg, Animesh and Yu, Guangwei},
  title     = {X-pool: Cross-modal language-video attention for text-video retrieval},
  booktitle = {Proceedings of the IEEE/CVF conference on computer vision and pattern recognition},
  pages     = {5006-5015},
  year      = {2022},
  type      = {Conference Proceedings}
}

@inproceedings{hur2025narrating,
  author    = {Hur, Chan and Hong, Jeong-hun and Lee, Dong-hun and Kang, Dabin and Myeong, Semin and Park, Sang-hyo and Park, Hyeyoung},
  title     = {Narrating the Video: Boosting Text-Video Retrieval via Comprehensive Utilization of Frame-Level Captions},
  booktitle = {Proceedings of the Computer Vision and Pattern Recognition Conference},
  pages     = {24077-24086},
  year      = {2025},
  type      = {Conference Proceedings}
}

@inproceedings{jeong2025learning,
  author    = {Jeong, Boseung and Park, Jicheol and Kim, Sungyeon and Kwak, Suha},
  title     = {Learning Audio-guided Video Representation with Gated Attention for Video-Text Retrieval},
  booktitle = {Proceedings of the Computer Vision and Pattern Recognition Conference},
  pages     = {26202-26211},
  year      = {2025},
  type      = {Conference Proceedings}
}

@inproceedings{jin2023diffusionret,
  author    = {Jin, Peng and Li, Hao and Cheng, Zesen and Li, Kehan and Ji, Xiangyang and Liu, Chang and Yuan, Li and Chen, Jie},
  title     = {Diffusionret: Generative text-video retrieval with diffusion model},
  booktitle = {Proceedings of the IEEE/CVF international conference on computer vision},
  pages     = {2470-2481},
  year      = {2023},
  type      = {Conference Proceedings}
}

@article{luo2022clip4clip,
  author  = {Luo, Huaishao and Ji, Lei and Zhong, Ming and Chen, Yang and Lei, Wen and Duan, Nan and Li, Tianrui},
  title   = {Clip4clip: An empirical study of clip for end to end video clip retrieval and captioning},
  journal = {Neurocomputing},
  volume  = {508},
  pages   = {293-304},
  issn    = {0925-2312},
  year    = {2022},
  type    = {Journal Article}
}

@inproceedings{wang2024text,
  author    = {Wang, Jiamian and Sun, Guohao and Wang, Pichao and Liu, Dongfang and Dianat, Sohail and Rabbani, Majid and Rao, Raghuveer and Tao, Zhiqiang},
  title     = {Text is mass: Modeling as stochastic embedding for text-video retrieval},
  booktitle = {Proceedings of the IEEE/CVF conference on computer vision and pattern recognition},
  pages     = {16551-16560},
  year      = {2024},
  type      = {Conference Proceedings}
}

@article{li2023dits,
  author  = {Wang, Jiamian and Wang, Pichao and Liu, Dongfang and Guan, Qiang and Dianat, Sohail and Rabbani, Majid and Rao, Raghuveer and Tao, Zhiqiang},
  title   = {Diffusion-inspired truncated sampler for text-video retrieval},
  journal = {Advances in Neural Information Processing Systems},
  volume  = {37},
  pages   = {3882-3906},
  year    = {2024},
  type    = {Journal Article}
}

@inproceedings{wang2023unified,
  author    = {Wang, Ziyang and Sung, Yi-Lin and Cheng, Feng and Bertasius, Gedas and Bansal, Mohit},
  title     = {Unified coarse-to-fine alignment for video-text retrieval},
  booktitle = {Proceedings of the IEEE/CVF international conference on computer vision},
  pages     = {2816-2827},
  year      = {2023},
  type      = {Conference Proceedings}
}

@inproceedings{wu2023cap4video,
  author    = {Wu, Wenhao and Luo, Haipeng and Fang, Bo and Wang, Jingdong and Ouyang, Wanli},
  title     = {Cap4video: What can auxiliary captions do for text-video retrieval?},
  booktitle = {Proceedings of the IEEE/CVF conference on computer vision and pattern recognition},
  pages     = {10704-10713},
  year      = {2023},
  type      = {Conference Proceedings}
}

@inproceedings{xu2016msrvtt,
  author    = {Xu, Jun and Mei, Tao and Yao, Ting and Rui, Yong},
  title     = {Msr-vtt: A large video description dataset for bridging video and language},
  booktitle = {Proceedings of the IEEE conference on computer vision and pattern recognition},
  pages     = {5288-5296},
  year      = {2016},
  type      = {Conference Proceedings}
}

@inproceedings{chen2011collecting,
  title     = {Collecting highly parallel data for paraphrase evaluation},
  author    = {Chen, David and Dolan, William B},
  booktitle = {Proceedings of the 49th annual meeting of the association for computational linguistics: human language technologies},
  pages     = {190--200},
  year      = {2011}
}

@inproceedings{wang2019vatex,
  title     = {Vatex: A large-scale, high-quality multilingual dataset for video-and-language research},
  author    = {Wang, Xin and Wu, Jiawei and Chen, Junkun and Li, Lei and Wang, Yuan-Fang and Wang, William Yang},
  booktitle = {Proceedings of the IEEE/CVF international conference on computer vision},
  pages     = {4581--4591},
  year      = {2019}
}

@inproceedings{chen2020fine,
  title     = {Fine-grained video-text retrieval with hierarchical graph reasoning},
  author    = {Chen, Shizhe and Zhao, Yida and Jin, Qin and Wu, Qi},
  booktitle = {Proceedings of the IEEE/CVF conference on computer vision and pattern recognition},
  pages     = {10638--10647},
  year      = {2020}
}

@inproceedings{radford2021learning,
  title        = {Learning transferable visual models from natural language supervision},
  author       = {Radford, Alec and Kim, Jong Wook and Hallacy, Chris and Ramesh, Aditya and Goh, Gabriel and Agarwal, Sandhini and Sastry, Girish and Askell, Amanda and Mishkin, Pamela and Clark, Jack and others},
  booktitle    = {International conference on machine learning},
  pages        = {8748--8763},
  year         = {2021},
  organization = {PmLR}
}

@article{loshchilov2017decoupled,
  title   = {Decoupled weight decay regularization},
  author  = {Loshchilov, Ilya and Hutter, Frank},
  journal = {arXiv preprint arXiv:1711.05101},
  year    = {2017}
}

@article{ho2020denoising,
  title   = {Denoising diffusion probabilistic models},
  author  = {Ho, Jonathan and Jain, Ajay and Abbeel, Pieter},
  journal = {Advances in neural information processing systems},
  volume  = {33},
  pages   = {6840--6851},
  year    = {2020}
}

@inproceedings{mithun2018learning,
  title     = {Learning joint embedding with multimodal cues for cross-modal video-text retrieval},
  author    = {Mithun, Niluthpol Chowdhury and Li, Juncheng and Metze, Florian and Roy-Chowdhury, Amit K},
  booktitle = {Proceedings of the 2018 ACM on international conference on multimedia retrieval},
  pages     = {19--27},
  year      = {2018}
}

@inproceedings{miech2019howto100m,
  title     = {Howto100m: Learning a text-video embedding by watching hundred million narrated video clips},
  author    = {Miech, Antoine and Zhukov, Dimitri and Alayrac, Jean-Baptiste and Tapaswi, Makarand and Laptev, Ivan and Sivic, Josef},
  booktitle = {Proceedings of the IEEE/CVF international conference on computer vision},
  pages     = {2630--2640},
  year      = {2019}
}

@article{patrick2020support,
  title   = {Support-set bottlenecks for video-text representation learning},
  author  = {Patrick, Mandela and Huang, Po-Yao and Asano, Yuki and Metze, Florian and Hauptmann, Alexander and Henriques, Joao and Vedaldi, Andrea},
  journal = {arXiv preprint arXiv:2010.02824},
  year    = {2020}
}

@article{lee2020parameter,
  title   = {Parameter efficient multimodal transformers for video representation learning},
  author  = {Lee, Sangho and Yu, Youngjae and Kim, Gunhee and Breuel, Thomas and Kautz, Jan and Song, Yale},
  journal = {arXiv preprint arXiv:2012.04124},
  year    = {2020}
}

@article{dhariwal2021diffusion,
  title   = {Diffusion models beat gans on image synthesis},
  author  = {Dhariwal, Prafulla and Nichol, Alexander},
  journal = {Advances in neural information processing systems},
  volume  = {34},
  pages   = {8780--8794},
  year    = {2021}
}

@article{song2020score,
  title   = {Score-based generative modeling through stochastic differential equations},
  author  = {Song, Yang and Sohl-Dickstein, Jascha and Kingma, Diederik P and Kumar, Abhishek and Ermon, Stefano and Poole, Ben},
  journal = {arXiv preprint arXiv:2011.13456},
  year    = {2020}
}

@article{ramesh2022hierarchical,
  title   = {Hierarchical text-conditional image generation with clip latents},
  author  = {Ramesh, Aditya and Dhariwal, Prafulla and Nichol, Alex and Chu, Casey and Chen, Mark},
  journal = {arXiv preprint arXiv:2204.06125},
  volume  = {1},
  number  = {2},
  pages   = {3},
  year    = {2022}
}

@article{saharia2022photorealistic,
  title   = {Photorealistic text-to-image diffusion models with deep language understanding},
  author  = {Saharia, Chitwan and Chan, William and Saxena, Saurabh and Li, Lala and Whang, Jay and Denton, Emily L and Ghasemipour, Kamyar and Gontijo Lopes, Raphael and Karagol Ayan, Burcu and Salimans, Tim and others},
  journal = {Advances in neural information processing systems},
  volume  = {35},
  pages   = {36479--36494},
  year    = {2022}
}

@article{ho2022video,
  title   = {Video diffusion models},
  author  = {Ho, Jonathan and Salimans, Tim and Gritsenko, Alexey and Chan, William and Norouzi, Mohammad and Fleet, David J},
  journal = {Advances in neural information processing systems},
  volume  = {35},
  pages   = {8633--8646},
  year    = {2022}
}

@article{singer2022make,
  title   = {Make-a-video: Text-to-video generation without text-video data},
  author  = {Singer, Uriel and Polyak, Adam and Hayes, Thomas and Yin, Xi and An, Jie and Zhang, Songyang and Hu, Qiyuan and Yang, Harry and Ashual, Oron and Gafni, Oran and others},
  journal = {arXiv preprint arXiv:2209.14792},
  year    = {2022}
}

@inproceedings{chun2021probabilistic,
  title     = {Probabilistic embeddings for cross-modal retrieval},
  author    = {Chun, Sanghyuk and Oh, Seong Joon and De Rezende, Rafael Sampaio and Kalantidis, Yannis and Larlus, Diane},
  booktitle = {Proceedings of the IEEE/CVF conference on computer vision and pattern recognition},
  pages     = {8415--8424},
  year      = {2021}
}

@article{chun2023improved,
  title   = {Improved probabilistic image-text representations},
  author  = {Chun, Sanghyuk},
  journal = {arXiv preprint arXiv:2305.18171},
  year    = {2023}
}

@inproceedings{reddy2025video,
  title={Video-colbert: Contextualized late interaction for text-to-video retrieval},
  author={Reddy, Arun and Martin, Alexander and Yang, Eugene and Yates, Andrew and Sanders, Kate and Murray, Kenton and Kriz, Reno and De Melo, Celso M and Van Durme, Benjamin and Chellappa, Rama},
  booktitle={Proceedings of the Computer Vision and Pattern Recognition Conference},
  pages={19691--19701},
  year={2025}
}

@inproceedings{lan2025hybrid,
  title={Hybrid-Tower: Fine-grained Pseudo-query Interaction and Generation for Text-to-Video Retrieval},
  author={Lan, Bangxiang and Xie, Ruobing and Zhao, Ruixiang and Sun, Xingwu and Kang, Zhanhui and Yang, Gang and Li, Xirong},
  booktitle={Proceedings of the IEEE/CVF International Conference on Computer Vision},
  pages={24497--24506},
  year={2025}
}

@article{hao2023uncertainty,
  title={Uncertainty-aware alignment network for cross-domain video-text retrieval},
  author={Hao, Xiaoshuai and Zhang, Wanqian},
  journal={Advances in Neural Information Processing Systems},
  volume={36},
  pages={38284--38296},
  year={2023}
}
\end{document}